\documentclass[letterpaper]{article} 
\usepackage{aaai25}  
\usepackage{times}  
\usepackage{helvet}  
\usepackage{courier}  
\usepackage[hyphens]{url}  
\usepackage{graphicx} 
\usepackage{natbib}  
\usepackage{caption} 
\usepackage{booktabs}       

\usepackage{makecell}
\usepackage{multirow}
\usepackage{algorithm}
\usepackage[noend]{algpseudocode}
\usepackage{amsmath}
\usepackage{amssymb}
\usepackage{mathtools}
\usepackage{amsthm}
\usepackage[capitalize]{cleveref}
\usepackage{caption}
\usepackage{subcaption}

\usepackage{xspace}
\usepackage{pifont}
\newcommand{\cmark}{\ding{51}}%
\newcommand{\xmark}{\ding{55}}%

\makeatletter
\DeclareRobustCommand\onedot{\futurelet\@let@token\@onedot}

\def\@onedot{\ifx\@let@token.\else.\null\fi\xspace}

\def\eg{\emph{e.g}\onedot} 
\def\ie{\emph{i.e}\onedot}

\makeatother
\usepackage{newfloat}
\usepackage{listings}
\DeclareCaptionStyle{ruled}{labelfont=normalfont,labelsep=colon,strut=off} 
\floatstyle{ruled}
\newfloat{listing}{tb}{lst}{}
\floatname{listing}{Listing}
\title{Context-Aware Token Selection and Packing for Enhanced Vision Transformer}

\author {
    Tianyi Zhang\textsuperscript{\rm 1},
    Baoxin Li\textsuperscript{\rm 2},
    Jae-sun Seo\textsuperscript{\rm 3},
    Yu Cao\textsuperscript{\rm 1}
}
\affiliations {
    \textsuperscript{\rm 1}University of Minnesota\\
    \textsuperscript{\rm 2}Arizona State University\\
    \textsuperscript{\rm 3}Cornell Tech\\
    zhan9167@umn.edu, baoxin.li@asu.edu, js3528@cornell.edu, yucao@umn.edu
}

\usepackage{bibentry}

\begin{document}

\maketitle

\begin{abstract}
In recent years, the long-range attention mechanism of vision transformers has driven significant performance breakthroughs across various computer vision tasks. However, the traditional self-attention mechanism, which processes both informative and non-informative tokens, suffers from inefficiency and inaccuracies.
While sparse attention mechanisms have been introduced to mitigate these issues by pruning tokens involved in attention, they often lack context-awareness and intelligence. These mechanisms frequently apply a uniform token selection strategy across different inputs for batch training or optimize efficiency only for the inference stage.
To overcome these challenges, we propose a novel algorithm: Select and Pack Attention (SPA). SPA dynamically selects informative tokens using a low-cost gating layer supervised by selection labels and packs these tokens into new batches, enabling a variable number of tokens to be used in parallelized GPU batch training and inference. Extensive experiments across diverse datasets and computer vision tasks demonstrate that SPA delivers superior performance and efficiency, including a 0.6 mAP improvement in object detection and a 16.4\% reduction in computational costs.
\end{abstract}

%

\section{Introduction} \label{intro section}
\begin{figure*}
    \centering
    \includegraphics[width=0.95\textwidth]{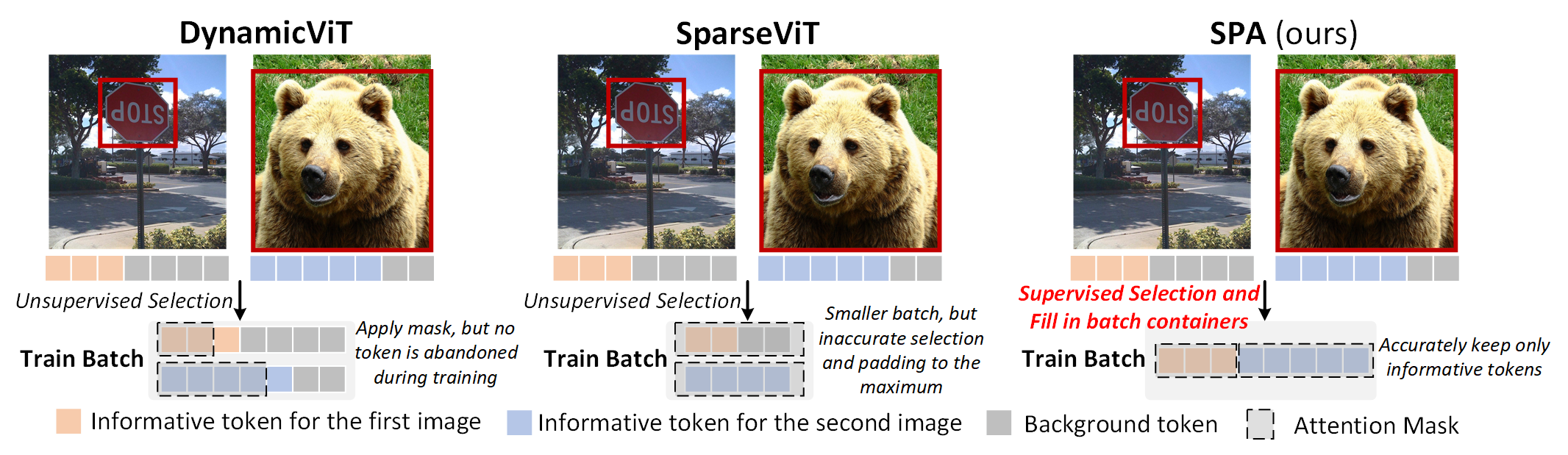}
    \caption{Previous sparse attention methods either reduce computation only during the inference stage or require padding the length of selected tokens to the maximum within a batch, which inevitably introduces background tokens. This leads to reduced efficiency and worse accuracy compared to our SPA.}
    \label{fig:intro}
\end{figure*}

Recent advancements in computer vision tasks such as image classification, segmentation, and object detection have seen Vision Transformers (ViTs) surpass traditional convolutional approaches \cite{dosovitskiy2020image, liu2021swin, xia2022vision, chen2023sparsevit} due to their powerful self-attention mechanisms. ViTs are particularly effective at capturing long-range dependencies, enabling the learning of global features that are crucial for complex visual understanding. However, this strength comes with a significant drawback: the computational overhead increases quadratically with the number of tokens \cite{liu2021swin, hua2022transformer}, leading to excessive and often unnecessary computations among irrelevant tokens. This not only raises the computational burden but also risks degrading performance by incorporating extraneous, often redundant, information in typical computer vision tasks.
As illustrated in \cref{fig:intro}, the use of self-attention in ViTs inadvertently processes a large amount of superfluous data, exacerbating computational inefficiency and potentially degrading task performance by introducing irrelevant information into the model's learning process. This issue is particularly severe when dealing with sparse data, where most pixels are not informative.

Numerous approaches have been proposed to address this issue by performing self-attention only on the most informative tokens. However, these methods still encounter significant challenges in both efficiency and performance.
\begin{itemize}
    \item Efficiency: The constraints of GPU batch training, where images within a batch often contain non-uniform numbers of informative tokens, pose challenges to parallelizing computation effectively. Some methods, such as SparseViT \cite{chen2023sparsevit}, address this by padding all effective tokens to match the maximum number in the batch, leading to inefficiencies, as illustrated in \cref{fig:intro}. Other approaches, like DynamicViT \cite{rao2021dynamicvit} and EViT \cite{liang2022not}, reduce computation only during inference by discarding a fixed number of tokens. However, these methods still attend to all tokens during training, employing an attention mask to focus on informative tokens, which, along with the mask prediction module, can result in training costs that exceed those of a standard ViT. The Deformable Attention Transformer (DAT) \cite{xia2022vision}, inspired by Deformable Convolutional Networks (DCN) \cite{dai2017deformable}, merely reduces the receptive field of query tokens while still computing all tokens, yielding minimal improvements in computational efficiency.
    \item Performance: Existing methods demonstrate effectiveness primarily in simpler tasks like image classification, where some degree of information loss is tolerable. However, their performance degrades in more complex tasks, such as object detection, which demand richer semantic information. For example, DynamicSwin \cite{rao2023dynamic}, a Swin-based DynamicViT, struggles in these scenarios due to inaccurate token selection, leading to significant information loss.
\end{itemize}

To address these challenges, we propose a novel Select and Pack Attention (SPA) mechanism that dynamically selects varying numbers of informative tokens from input batches, supervised by selection labels, and packs them into new batches for parallelized training. Specifically, we introduce a linear gating layer to generate scores for token selection, supervised by a multi-scale selection label derived from object labels (e.g., bounding boxes, instance segmentation labels). After selection, the chosen tokens are placed into uniform-sized package containers to form new batches, as illustrated in \cref{fig:intro}. For attention computation within each container, tokens attend only to those from the same original image, ignoring tokens from other images by using attention masks.
Additionally, SPA can be effectively integrated with the window-based attention proposed by Swin Transformer \cite{liu2021swin}, benefiting from the window shifting operation that captures information across windows. To prevent information loss across package containers, we shift the feature maps every two transformer blocks, ensuring that token pairs placed into containers vary, allowing the attention computation to encompass all tokens.
Based on SPA, we propose a backbone network, Select and Pack Transformer (SPT), featuring a hierarchical architecture to generate image representations at various scales for downstream computer vision tasks.  Similar to DAT \cite{xia2022vision}, to avoid mis-selection at the early stage which may cause serious information loss, we leverage our SPA from the third stage with the adapted image features. 
Ultimately, SPA addresses the efficiency issue by selecting only informative tokens and packing them into new batches, enabling efficient parallelized computation for both training and inference. Moreover, by leveraging selection label supervision, SPA improves performance in complex computer vision tasks, such as object detection.
Comprehensive experiments on four well-known datasets demonstrate the efficacy of SPA across multiple computer vision tasks.

To summarize, our main contributions are as follows:
\begin{itemize}
    \item We propose a novel sparse attention mechanism, Select and Pack Attention (SPA), to enhance both the efficiency and performance of Vision Transformers. For efficiency, SPA dynamically selects informative tokens from images in a batch using a linear gating layer and packs them together to enable efficient GPU batch training and inference. For performance, we introduce a multi-scale selection label to explicitly supervise token selection, thereby outperforming existing methods even in complex computer vision tasks.
    \item By effectively integrating our SPA mechanism with Swin blocks, which use a window shifting trick to capture information across packages, we propose a backbone network with a hierarchical structure called the Select and Pack Transformer (SPT). SPT can generate features at various scales, making it suitable for many computer vision tasks.
    \item Through extensive experiments on four diverse datasets, we demonstrate the superior performance of our Select and Pack Transformer (SPT) across a range of computer vision tasks. SPT consistently outperforms state-of-the-art methods with a 0.6 mAP improvement in object detection, a 0.24 mAP increase in multi-label classification, a 7.05 boost in top-1 accuracy for image classification, and a 16.4\% reduction in computation cost.
\end{itemize}

\section{Related Work}\label{sec:related}
\subsection{Transformer in Computer Vision}
Given the remarkable success of transformers in natural language processing (NLP), this architectural paradigm is progressively permeating diverse computer vision tasks \cite{vaswani2017attention, bao2021beit, touvron2021training, he2022masked, zhang2024patch, konstantinidis2023multi, yang2022retargeting, kang2022tie, ni2024earnings, ni2024timeseries, zhou2024reconstruction, fan2024advanced, fan2024towards}. For instance, Vision Transformer (ViT) divides input images into $16\times16$ patches, which are subsequently treated as tokens for the application of the attention mechanism \cite{dosovitskiy2020image}.
In image segmentation, SAM \cite{kirillov2023segment} introduces a prompt-based algorithm, setting new benchmarks across state-of-the-art methods. For object detection (OD), DETR conceptualizes it as a direct set prediction problem and designs a transformer-based network \cite{carion2020end}. DINO advances self-supervised learning to propose a novel network rooted in the ViT architecture \cite{caron2021emerging}. Additionally, in the domain of super-resolution, transformers such as SwinIR have demonstrated exceptional capability in capturing long-range dependencies for improved visual representations \cite{liang2021swinir, zhang2023transformer}.

\subsection{Efficient Transformers}
Despite their advantages in global feature extraction via self-attention across all tokens, Vision Transformers (ViTs) are hindered by significant computational overhead. This overhead primarily arises because the computation of attention weights scales quadratically with the number of tokens. To address this challenge, two main approaches have been proposed: pruning the number of tokens for sparse attention or developing mechanisms with linear complexity.
For sparse attention, Swin Transformer \cite{liu2021swin} introduces window-based and shifted window-based self-attention mechanisms, significantly reducing computational demands within localized windows. MAE \cite{he2022masked} employs random masking to decrease token computation. Sparse Transformer \cite{child2019generating} proposes two new attention mechanisms to limit the number of tokens in each attention computation. Besides these data-agnostic sparse attention methods, DynamicViT \cite{rao2021dynamicvit} proposes a dynamic token sparsification framework to prune redundant tokens progressively and dynamically based on the input, SparseViT \cite{chen2023sparsevit} optimizes computation by selecting tokens based on the $l_2$ norm of window activations, prioritizing features with higher scores. Inspired by Deformable Convolutional Networks (DCNs), DAT \cite{xia2022vision} employs an offset network to refine the query token's receptive field, further enhancing computational efficiency. Our Select and Pack Attention (SPA) mechanism also belongs to this category.
For linear transformers, Transformer-VQ \cite{lingle2023transformer} achieves efficient attention with linear complexity using vector-quantized keys and a novel caching mechanism. FLASH attention \cite{hua2022transformer} proposes a new transformer with linear time complexity, utilizing Gated Linear Units (GLUs) \cite{shazeer2020glu}.

\section{Methodology}\label{sec:method}
\subsection{Overall Architecture}
\begin{figure*}[hbt]
    \centering
    \includegraphics[width=0.95\textwidth]{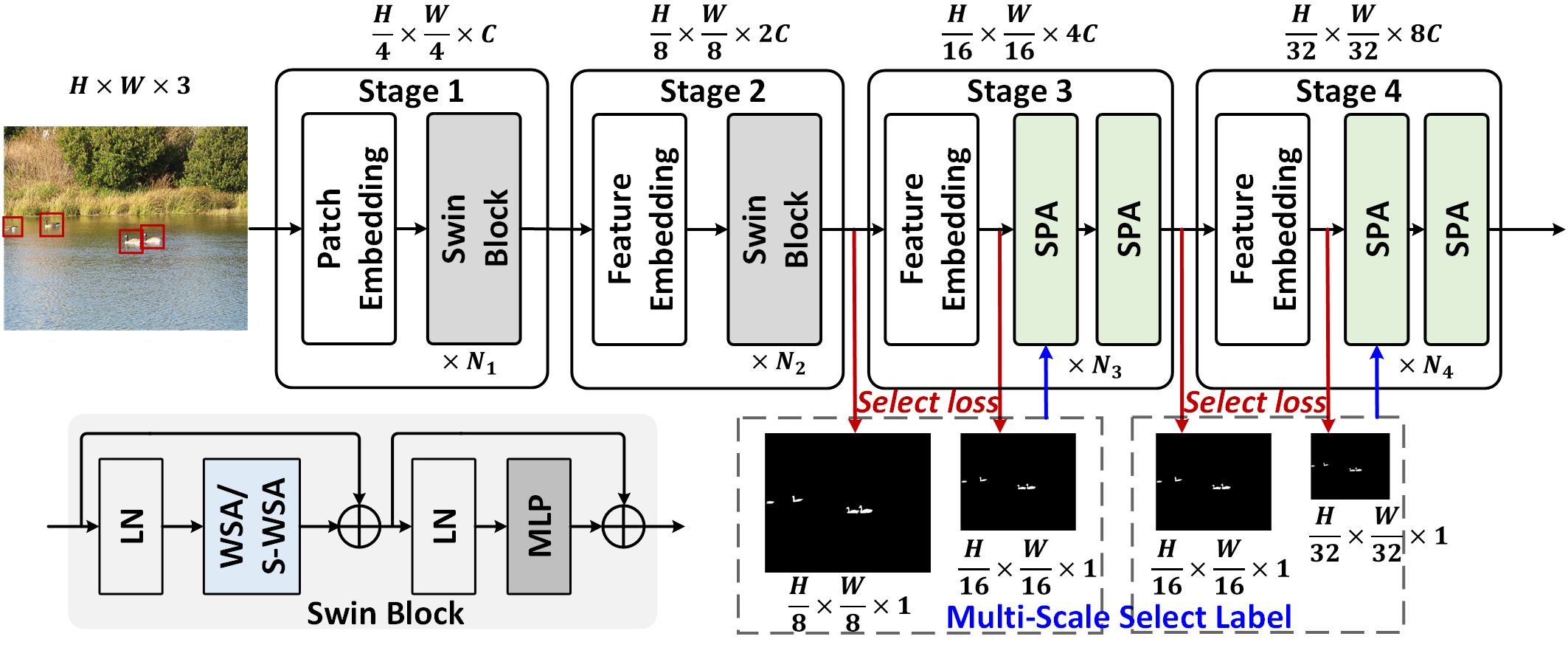}
    \caption{Overall architecture of Select and Packing Transformer (SPT). The hierarchical structure can generate features with various scales as common backbone networks. The SPA blocks in the last two stages can improve both efficiency and accuracy by disregarding uninformative tokens.}
    \label{fig:chart}
\end{figure*}

As illustrated in \cref{fig:chart}, our Select and Pack Transformer (SPT) features a hierarchical structure composed of four stages. Each stage generates image representations of varying sizes, resulting in a total of four different scales of representation.

\begin{figure}[htb]
     \centering
     \begin{subfigure}[b]{0.42\textwidth}
         \centering
         \includegraphics[width=\textwidth]{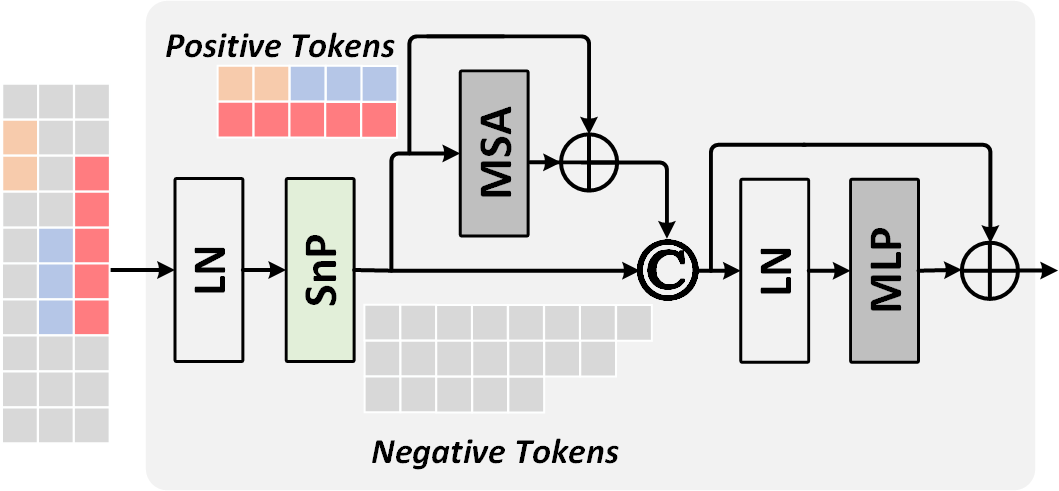}
         \caption{SPA block.}
         \label{fig:spa1}
     \end{subfigure}
     \hfill
     \begin{subfigure}[b]{0.42\textwidth}
         \centering
         \includegraphics[width=\textwidth]{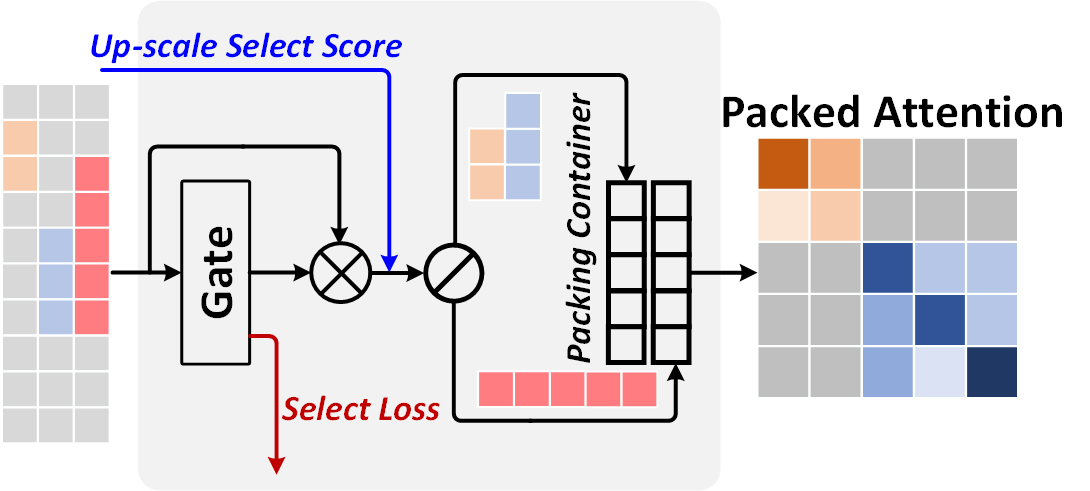}
         \caption{SnP block.}
         \label{fig:spa2}
     \end{subfigure}
        \caption{(a) Our SPA computes attention only for informative tokens. (b) Our SnP block selects informative tokens under multi-scale supervision and packs selected tokens for batch training and inference. The packed tokens attend to only tokens from the same image.}
        \label{fig:spa}
\end{figure}

Specifically, suppose we consider a small $4\times4$ patch as a single token, the input image $\boldsymbol{x}\in\mathbb{R}^{H\times W\times 3}$ ($H$ and $W$ are the input image height and width), are progressively embedded into representations $\boldsymbol{r}_1\in\mathbb{R}^{\frac{H}{4}\times\frac{W}{4} \times C}$,  $\boldsymbol{r}_2\in\mathbb{R}^{\frac{H}{8}\times\frac{W}{8} \times 2C}$,  $\boldsymbol{r}_3\in\mathbb{R}^{\frac{H}{16}\times\frac{W}{16} \times 4C}$,  $\boldsymbol{r}_4\in\mathbb{R}^{\frac{H}{32}\times\frac{W}{32} \times 8C}$ ($C$ is the embedding dimension of the first patch embedding layer) stage by stage. Each stage is structured around an embedding block for feature map downsampling, followed by $N_i$ transformer blocks tasked with feature learning ($N_i$ signifies the block count in the $i$th stage).
Similar to Swin \cite{liu2021swin, liu2022swin}, the embedding block of the first stage $f_{\theta_1}$ employs a convolution layer while the subsequent embedding block consists of a patch merging layer that concatenates features as groups of $2\times2$ patches and a linear layer for feature projection.
For the transformer blocks, the first two stages $f_{\theta_1}$, $f_{\theta_2}$ utilize standard Swin Transformer blocks, whereas the latter two stages $f_{\theta_3}$, $f_{\theta_4}$ incorporate our Select and Pack Attention (SPA) block. This design decision is informed by observations from DAT \cite{xia2022vision}, which noted that early-stage transformer block replacement diminishes accuracy due to the model's inability to efficiently distinguish positive tokens based on shallow features.
Our SPA blocks in the second and third stages not only generate outputs for subsequent layers but also transfer the score map to the next stage for  the multi-scale supervision $\boldsymbol{s}_0 \in \mathbb{R}^{\frac{H}{8}\times \frac{W}{8}\times 1}$ and $\boldsymbol{s}_1 \in \mathbb{R}^{\frac{H}{16}\times \frac{W}{16}\times 1}$ for computing select loss. With the selection map $\boldsymbol{s}_2 \in \mathbb{R}^{\frac{H}{32}\times \frac{W}{32}\times 1}$ generated in the last stage, there are a total of three different scales. The complete process is as follows:

\begin{align}
    \boldsymbol{r}_1 = f_{\theta_1}(\boldsymbol{x}),  \;
    \boldsymbol{r}_2 &= f_{\theta_2}(\boldsymbol{r}_1), \;
    \boldsymbol{s}_0 = f_{\theta_g}(\boldsymbol{r}_2) \\
    \boldsymbol{r}_3, \boldsymbol{s}_1 &= f_{\theta_3}(\boldsymbol{r}_2, \boldsymbol{s}_0), \\
    \boldsymbol{r}_4, \boldsymbol{s}_2 &= f_{\theta_4}(\boldsymbol{r}_3, \boldsymbol{s}_1),
\end{align}
where $\boldsymbol{r}_1$, $\boldsymbol{r}_2$, $\boldsymbol{r}_3$, and $\boldsymbol{r}_4$ are output representations of four stages. $f_{\theta_1}$, $f_{\theta_2}$, $f_{\theta_3}$, and $f_{\theta_4}$ denote the four stage models. And $\boldsymbol{s}_0$, $\boldsymbol{s}_1$  and $\boldsymbol{s}_2$ denote the predicted score map for selection from the last three stages, separately. $f_{\theta_g}$ is the gating layer to generate scores for the output of stage 2.


\subsection{Select and Pack Attention (SPA)}

Inspired by the gated networks in Mixture of Experts (MoE) \cite{petersen2022deep, huang2020gatenet, shazeer2017outrageously, aoki2022heterogeneous, chen2022towards} and heterogeneous federated learning \cite{lin2021metagater, ye2023heterogeneous}, which adeptly guide models in selecting appropriate computational paths and enhancing task-specific generalization, we design a Select and Pack (SnP) block. This block utilizes a linear gating layer to select informative tokens and pack them into fixed-size package containers, generating new batches for GPU training or inference. While positive tokens undergo multi-head self-attention (MSA), negative tokens are directly passed to the feedforward network, as illustrated in \cref{fig:spa1}.

\textbf{Multi-Scale Supervised Selection.} 
Although token selection can be implicitly guided by the final objective, our experiments reveal that the gating layer tends to assign large values to all tokens, leading to the selection of too many tokens and reduced efficiency. To address this, we introduce a selection label based on object labels, which directly indicate areas of interest, such as instance segmentation masks or object detection bounding boxes. For segmentation, a binary mask assigns a value of 1 to all object pixels and 0 otherwise. For object detection, an aggregated binary mask is formed by stacking all bounding boxes.
However, a single-scale label overly restricts token selection, causing significant information loss and poor performance. To mitigate this, we reduce the Gumbel-Softmax function's threshold and integrate multi-scale select labels. As shown in \cref{fig:spa2}, each SPA block in SPT not only uses the selection scale matching the representation but also incorporates scores from up-scaled features, adjusted via max-pooling to match the correct feature size. This approach selects the maximum scores from two scales to include more informative tokens, thereby enhancing performance.

Specifically, given flattened input batch $\boldsymbol{r}\in\mathbb{R}^{B\times N \times C}$ ($B$, $N$ and $C$ are the batch size, the length of each image representation and the number of channels, separately), the gate $f_{\theta_g}$ assigns scores $\boldsymbol{s}\in\mathbb{R}^{B\times N \times 1}$ to each token. Then,  we element-wise multiply the normalized scores by a sigmoid layer with the input representations to obtain the gated representations $\boldsymbol{r}_g \in \mathbb{R}^{B\times N \times C}$. After that, we leverage the Gumbel-Softmax function \cite{jang2016categorical} to separate positive tokens (\ie informative tokens), $\boldsymbol{r}_p\in\mathbb{R}^{N_p\times C}$ ($N_p$ is the number of positive tokens from all images in the batch).
The procedure unfolds as follows:
\begin{align}
    \boldsymbol{s} &= \text{Max}(f_{\theta_g}(\boldsymbol{r}), \boldsymbol{s}_{up}), \\
    \boldsymbol{r}_g &= \text{Sigmoid}(\boldsymbol{s})\odot\boldsymbol{r}, \\
    \boldsymbol{r}_p &= \text{Gumbel-Softmax}(\boldsymbol{s})\odot\boldsymbol{r}_g
\end{align}
where $\boldsymbol{r}$, $\boldsymbol{s}_{up}$, $\boldsymbol{s}$, $\boldsymbol{r}_g$, $\boldsymbol{r}_p$ denotes the input representation, scores for up-scale features, scores for this scale, gated representation, and output positive tokens, separately. And $\odot$ is element-wise multiplication with boardcasting. $f_{\theta_g}$ is the linear gating layer.

\textbf{Token Packing.}
After the dynamic selection for each input image, the lengths of selected tokens vary. To avoid padding all tokens to the maximum length, which would introduce significant computational overhead, we pack the selected tokens into new batches. Inspired by \cite{dehghani2024patch}, we set a series of package containers with a fixed length $L$ and fill them with the selected tokens.
After packing all selected tokens, if the total number of tokens is not a multiple of the packing length, we only pad the last package. This approach is significantly more efficient than padding the selected tokens for all images in the batch. Consequently, we obtain packed tokens, $\boldsymbol{p}\in \mathbb{R}^{B'\times L \times C}$ ($B'$ is the batch size of packed tokens), and the number of tokens is much smaller than the original input, especially for sparse data. And the attention computation depends on $L$, similar to the window size $M$ of Swin. And we set $L$ to be $M^2$. Specifically, for input representation batch $\boldsymbol{r}\in\mathbb{R}^{B\times N \times C}$, the complexity of regular multi-head self-attention (MSA), window-based multi-head self-attention (W-MSA), and our SPA are as follows:
\begin{align}
    \Omega(\text{MSA}) &= B(4NC^2+2N^2C), \\
    \Omega(\text{W-MSA}) &= B(4NC^2+2M^2NC), \\
    \Omega(\text{SPA}) &= B(NC+NC^2)+B'(3LC^2+2L^2C),
\end{align}
Compared to MSA, W-MSA is more efficient since the complexity is linear to the original token length $N$. However, our SPA is not only linear to $N$, the new batch size $B'$ is also much smaller than $B$, resulting in higher efficiency.
Additionally, for the self-attention of the packed tokens, we employ an attention mask to ensure that all tokens attend only to tokens from the same image, as illustrated in \cref{fig:spa2}.

\subsection{Loss Function}
The loss function of our SPT Transformer comprises the loss for the target task and the selection loss. For the selection loss (Details in Appendix A), we adopt binary cross-entropy and sum over all SPA blocks as follows,
\begin{equation}
    \mathcal{L}_{select} = -\sum_{block} (\boldsymbol{y} \log \boldsymbol{s}+(1-\boldsymbol{y}) \log (1-\boldsymbol{s}))
\end{equation}
where $\boldsymbol{s}$ is the normalized score map by Sigmoid layer, and $\boldsymbol{y}$ is the ground truth label. 

The loss function of our SPT is $\mathcal{L}_{SPT} = \mathcal{L}_{task} + \alpha\mathcal{L}_{select}$, where $\alpha$ is hyperparameter to adjust the weights of losses. 
And we summarize our algorithm in Appendix B.

\section{Experimental Results}\label{sec:experiment}

\subsection{Data and Experimental Setup}

\begin{table*}[hbt]
\centering
\begin{tabular}{@{} l|c c c c c c |c  ccc@{}}
\toprule
\multirow{2}{*}{\textbf{Methods}} & \multicolumn{6}{c|}{\textbf{OD Performance}}  & \multirow{2}{*}{\textbf{\makecell{\#Params\\(M)}}} & \multirow{2}{*}{\textbf{\makecell{FLOPs\\(G)}}} & \multirow{2}{*}{\textbf{\makecell{FPS\\(image/s)}}}\\ 
\cmidrule(){2-7}
             &  $AP$   &  $AP_{50}$ & $AP_{75}$ & $AP_{S}$ & $AP_{M}$ & $AP_{L}$ & 
                  \\ \midrule
     Swin-T \cite{liu2021swin} &  22.4     &  34.6      &     24.6    &  7.9   & 20.1   &  46.8   & 48  & 267 & 50    \\
         DynamicSwin-T \cite{rao2023dynamic} & 22.0      &  33.1      &  23.7    &   9.2     &    19.7    &   44.7    & 48    & 272 & 46   \\
         \textbf{SPT-T} (ours) &  \textbf{22.6}      &    33.1     &    24.6     &  8.8     &      18.5   &    47.7   & 48  & \textbf{255} & \textbf{58} \\
        \midrule 
        Swin-S \cite{liu2021swin} &  22.6     &    34.9     &   24.9    &   8.1    &   20.3   &   47.1 &  69  & 359 & 32    \\
         DynamicSwin-S \cite{rao2023dynamic} &  22.3     &   33.4       &    24.2    &   9.0     &     19.9  &   45.4    & 69  & 363 & 32   \\
         \textbf{SPT-S} (ours) &  \textbf{22.9}      &   33.8     &   25.2     &  8.7     &      19.8   &    48.2    & 69  & \textbf{326} & \textbf{34} \\
        \midrule 
         Swin-B \cite{liu2021swin} &  22.7    &    35.1     &   25.1    &   8.2     &  20.5  &   47.6    & 107  & 508 & 18   \\
        DynamicSwin-B \cite{rao2023dynamic} &    22.5   &   33.6        &   24.6   &   8.5   &   20.2  &   46.3   & 107   & 517 & 18  \\
         \textbf{SPT-B} (ours) & \textbf{23.1}   &   34.3   &  25.5   &   8.2    &     20.6  &  48.6    & 107   & \textbf{432} & \textbf{20} \\
        \bottomrule
\end{tabular}
\caption{Our SPT-based Mask RCNN achieves better object detection (OD) performance with less total computation on BDD100K for all three configurations.}
\label{tab:od1}
\end{table*}

\begin{table*}[hbt]
\centering
\begin{tabular}{@{}l | l|c c c c c c |c  ccc@{}}
\toprule
\multirow{2}{*}{\textbf{Dataset}} &\multirow{2}{*}{\textbf{Methods}} & \multicolumn{6}{c|}{\textbf{OD Performance}}  & \multirow{2}{*}{\textbf{\makecell{\#Params\\(M)}}} & \multirow{2}{*}{\textbf{\makecell{FLOPs\\(G)}}} & \multirow{2}{*}{\textbf{\makecell{FPS\\(im/s)}}}\\ 
\cmidrule(){3-8}
           &    &  $AP$   &  $AP_{50}$ & $AP_{75}$ & $AP_{S}$ & $AP_{M}$ & $AP_{L}$ & 
                  \\ \midrule
     \multirow{6}{*}{\textbf{BDD-S}} 
        & Swin-T \cite{liu2021swin} &  5.5      &    8.6       &     5.9     &      1.4     &      2.7    &   15.4    & 48    & 267 & 50   \\
       &  DynamicSwin-T\cite{rao2023dynamic} &  4.7      &    8.4       &   3.7     &   1.0   &   2.6   &   12.8     & 48    & 272& 46  \\
       &  \textbf{SPT-T} (ours) &  \textbf{5.6}      &  9.0      & 6.4     &      1.5     &      2.7    &   15.4  & 48   & \textbf{251} & \textbf{62} \\
        \cmidrule(){2-11}
      &   Swin-S \cite{liu2021swin} &  5.4     &    9.0      &     6.0     &      1.7     &    2.9   &     14.1     & 69  & 359 & 32   \\
     &    DynamicSwin-S\cite{rao2023dynamic} &  5.2     &    8.2        &     6.0    &   0.9   &      2.3   &    14.3    & 69   & 363 & 32  \\
      &   \textbf{SPT-S} (ours) & \textbf{5.7}   &  9.2   & 6.6   &      1.7    &   2.9  &   15.5    & 69   & \textbf{320} & \textbf{35} \\
        \bottomrule
\end{tabular}
\caption{For the more challenging BDD-S dataset, which contains sparse data, our SPT significantly outperforms baseline models while requiring less computation.}
\label{tab:od2}
\end{table*}

\begin{table}[hbt]
\centering
\begin{tabular}{@{}l|l| c |c  |c@{}}
\toprule
\textbf{Dataset} & \textbf{Methods}  & \textbf{AP}  &  \textbf{\makecell{FLOPs\\(G)}} & \textbf{\makecell{FPS\\(im/s)}}
                  \\ \midrule
      \multirow{2}{*}{\textbf{COCO-S}} &  Swin-T &  2.5    & 267   & 50 \\
       &  \textbf{SPT-T} (ours) &  2.6 & 258 &  \textbf{59 } \\
        
        
    \bottomrule
\end{tabular}
\caption{On the COCO-S dataset, our SPT also performs exceptionally well.}

\label{tab:coco}
\end{table}

To demonstrate the efficacy of our SPT Transformer in complex computer vision tasks, we primarily conducted experiments on object detection, using the BDD100K and COCO2017 datasets. 
Beyond these standard datasets, we explored our method on sparse data, where token selection is more challenging, further validating the effectiveness of our approach.
We generated sparse datasets by selecting images with low object ratios. 
Specifically, we selected images from COCO2017 with object pixel ratios smaller than 20\%, creating the COCO-S dataset. Similarly, we selected images from BDD100K with object ratios smaller than 25\%, resulting in the BDD-S dataset.
Additionally, to further showcase the robustness of our SPT Transformer, we extended our experiments to a range of simpler computer vision tasks. 
We evaluated multi-label classification using the PASCAL VOC 2012 dataset, selecting images with object pixel ratios smaller than 25\% to create VOC-S. For image classification, since datasets for this task are typically dense, we instead padded the original images with black pixels to make them sparse. Specifically, we selected the Tiny ImageNet subset of ImageNet-1K, which contains 100,000 training images and 10,000 validation images, and provides object labels for selection supervision. We padded black background pixels to the original images to make them sparse, called IN-S. After the $2\times2$ padding,  the original images were positioned in the upper-left corner, while the remaining 75\% of the area was filled with black pixels.   

To evaluate OD, we utilize the Mask RCNN framework, and replace the backbone network with our SPT or other baselines. We adopt the default training settings, such as 36 max training epochs, batch size of 2. In addition, we set the threshold of Gumbel-Softmax to 0.01, and set the select loss weight $\alpha$ to 0.01.
Experiments were performed on two Linux servers, each outfitted with dual NVIDIA L40S GPUs. 

\textbf{Evaluation Metrics.}
For object detection and multi-label classification, we employ mAP as the evaluation metric. And for image classification, we adopt Top-1 accuracy. In addition, we evaluate the select ratios for each SPA block, which is the number of selected tokens over the total number.

\subsection{SPT for Object Detection}
In \cref{tab:od1}, we compare our SPT with other baselines for the tiny, small and base configurations of the Swin Transformer (i.e., Swin-T, Swin-S and Swin-B) on BDD100K.
The GFLOPs are computed over backbone, FPN and detection head with RGB input image at the resolution of $1280\times800$ for training stage. For a clearer comparison, we evaluate the throughput (i.e., FPS) only over the backbone network on a machine with an NVIDIA L40S GPU, as including other components would result in values that are too small. Under the same settings, our approach achieves the best performance with the lowest computation cost across all three configurations. Specifically, our SPT-B model improves object detection results on BDD100K from 22.5, achieved by DynamicSwin—an existing state-of-the-art sparse attention method—to 23.1, while reducing computation by 16.4\% for both training and inference.

\textbf{Performance on Sparse Data.}
When dealing with the more challenging BDD-S dataset, as shown in \cref{tab:od2}, the GFLOPs were reduced from 272 to 251 for SPT-T and from 363 to 320 for SPT-S, representing reductions of 7.72\% and 11.8\%, respectively. These models also achieved performance improvements of 19.1\% and 9.6\%, respectively. In experiments on COCO-S, our SPT also outperformed other methods while requiring less computation in \cref{tab:coco}. These results collectively demonstrate the superiority of our SPA in accurately selecting informative tokens under the supervision of multi-scale selection labels.
Additional results can be found in Appendix C.

\subsection{SPT for Other Computer Vision Tasks}
In addition to the complex object detection task, we also evaluated our SPT on simpler tasks, including multi-label classification and image classification.

\begin{table}[h]
\centering
\begin{tabular}{@{}l|l| c|c@{}}
\toprule
\textbf{Dataset} & \textbf{Methods}  & \textbf{Mean Select Ratio(\%)}  &  \textbf{mAP}
                  \\ \midrule
      \multirow{2}{*}{\textbf{VOC-S}} &  Swin &  100   & 44.36    \\
        &  \textbf{SPT} (ours) &  \textbf{29.6} & \textbf{44.60}    \\
   
    \bottomrule
\end{tabular}
\caption{The SPA block reduces the computation with a low select ratio and achieves better performance in multi-label classification on VOC-S.}
\label{tab:voc}
\end{table}

\begin{table}[hbt]
\centering
\begin{tabular}{@{}c|l| c |c@{}}
\toprule
\textbf{Dataset} & \textbf{Methods}  & \textbf{Mean Select Ratio(\%)}  &  \textbf{Acc.}
                  \\ \midrule
      \multirow{2}{*}{\textbf{IN-S}} &  Swin &  100  & 29.10   \\
        &  \textbf{SPT} (ours) &  22.96 & \textbf{32.75}    \\
        
    \midrule
      \multirow{2}{*}{\textbf{\makecell{Tiny\\IN-1K}}} &  Swin  &  100 &  36.12    \\
        &  \textbf{SPT} (ours) & 76.20 & \textbf{43.17 }     \\
        
    \bottomrule
\end{tabular}
\caption{Our SPT performs both accurately and efficiently for image classification tasks.}
\label{tab:in}
\end{table}

\begin{figure}[hbt]
    \centering
    \includegraphics[width=0.95\columnwidth]{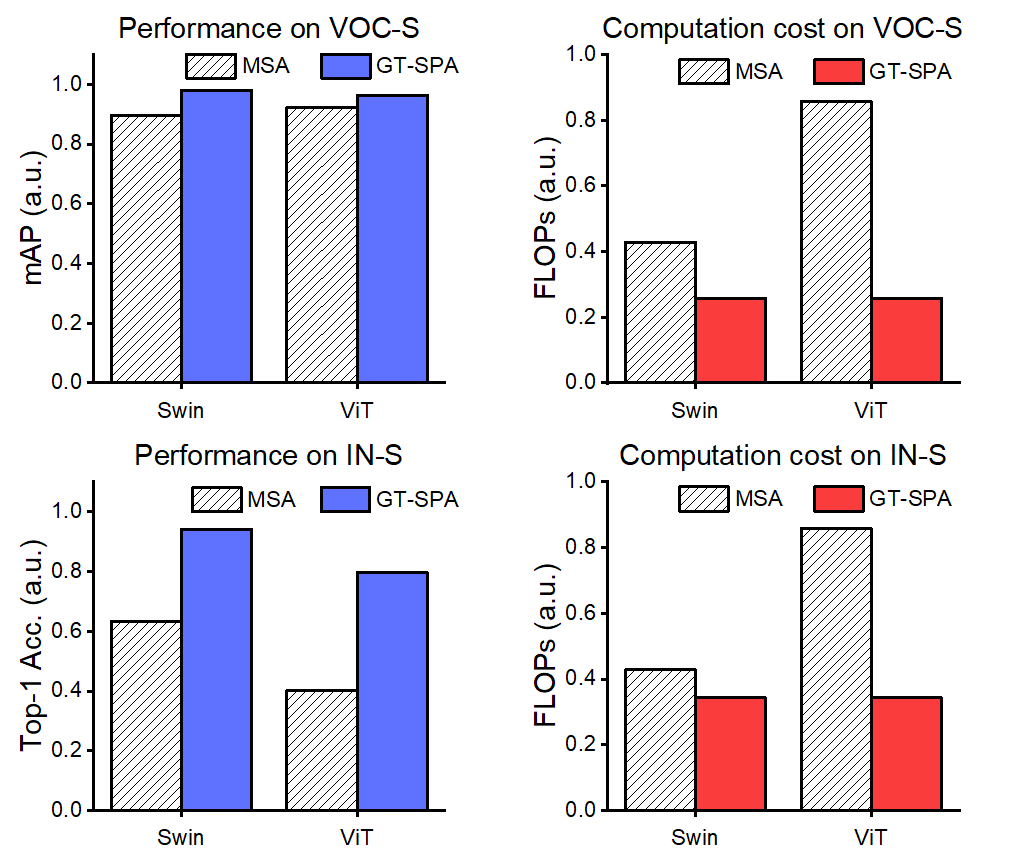}
    \caption{Under ground truth (GT) supervision, attending to only informative tokens can achieve better performance and efficiency.}
    \label{fig:gt}
\end{figure}

\begin{table*}[hbt]
\centering
\begin{tabular}{@{}c| c c c c |c c c c|c@{}}
\toprule
 \multirow{2}{*}{\textbf{\# SPA Blocks}} & \multicolumn{4}{c|}{\textbf{Select Ratios of stages (w/o $\mathcal{L}_{select}$)}}  & \multicolumn{4}{c|}{\textbf{ Select Ratios of stages (w/ $\mathcal{L}_{select}$) }}  &  \multirow{2}{*}{\textbf{Top-1 Acc.}} \\ 
\cmidrule(){2-9}
              & $1_{st}$   &  $2_{nd}$ & $3_{rd}$ & $4_{th}$ & $1_{st}$   &  $2_{nd}$ & $3_{rd}$ & $4_{th}$
                  \\ \midrule 
                  2 & \xmark      &    \xmark     &     \xmark    &      36.5   &     \xmark   &     \xmark    & \xmark  &25.02 & 31.81    \\
     4   &    \xmark      &   \xmark     &     85.02   &      52.41   &     \xmark   &     \xmark   & 23.85  & 25.0 & 31.92    \\
     6   &   \xmark     &    \xmark      &     92.56     &      88.42     &     \xmark    &   \xmark  & 23.11 & 25.01   &32.04 \\
  \textbf{ 8 }  &    \xmark      &   \xmark     &     93.21    &      81.74   &     \xmark   &     \xmark  & 22.68  & 25.01 & \textbf{32.75}    \\
   10   &    \xmark      &   99.24     &     82.15    &      71.45   &     \xmark   &     20.18   & 22.56  & 25.0 & 32.41    \\
   12   &    83.49      &   99.12     &     88.64    &      79.86   &     13.42   &     20.78   & 22.36  & 25.0 & 30.20    \\
    \bottomrule
\end{tabular}
\caption{ Experiments are on IN-S. In alignment with DAT, starting to replace the Swin blocks form the third stage works the best (\ie 8 SPA blocks totally). The selection at early stage will lead to information loss.}
\label{tab:blks}
\end{table*}

\textbf{Multi-Label Classification.}
For multi-label classification on VOC-S, Swin outperforms ViT, improving the mAP from 43.24 to 44.36. However, with our SPA, performance is further improved to 44.6 with a much lower computational cost. In \cref{tab:voc}, we show the mean of select ratios for SPA blocks (Detailed ratios are included in Appendix D). Overall, SPT reduces the GFLOPs for VOC-S by 10.2\%.

\textbf{Image Classification.}
On the original Tiny ImageNet dataset (Tiny IN-1K), as shown in \cref{tab:in}, the high selection ratios indicate minimal efficiency improvement, as this dataset is very dense. However, we observed an increase in Top-1 accuracy from 36.12 to 43.17, further demonstrating the effectiveness of our proposed attention mechanism in focusing on informative tokens.

For the more challenging IN-S dataset, our SPA selects approximately 23\% of the tokens for attention computation, aligning with the ground truth object pixel ratio. This approach not only improves Top-1 accuracy but also achieves a 10.5\% reduction in computation by disregarding background information, as shown in \cref{tab:in}.

\begin{figure}[hbt]
    \centering
    \includegraphics[width=0.98\columnwidth]{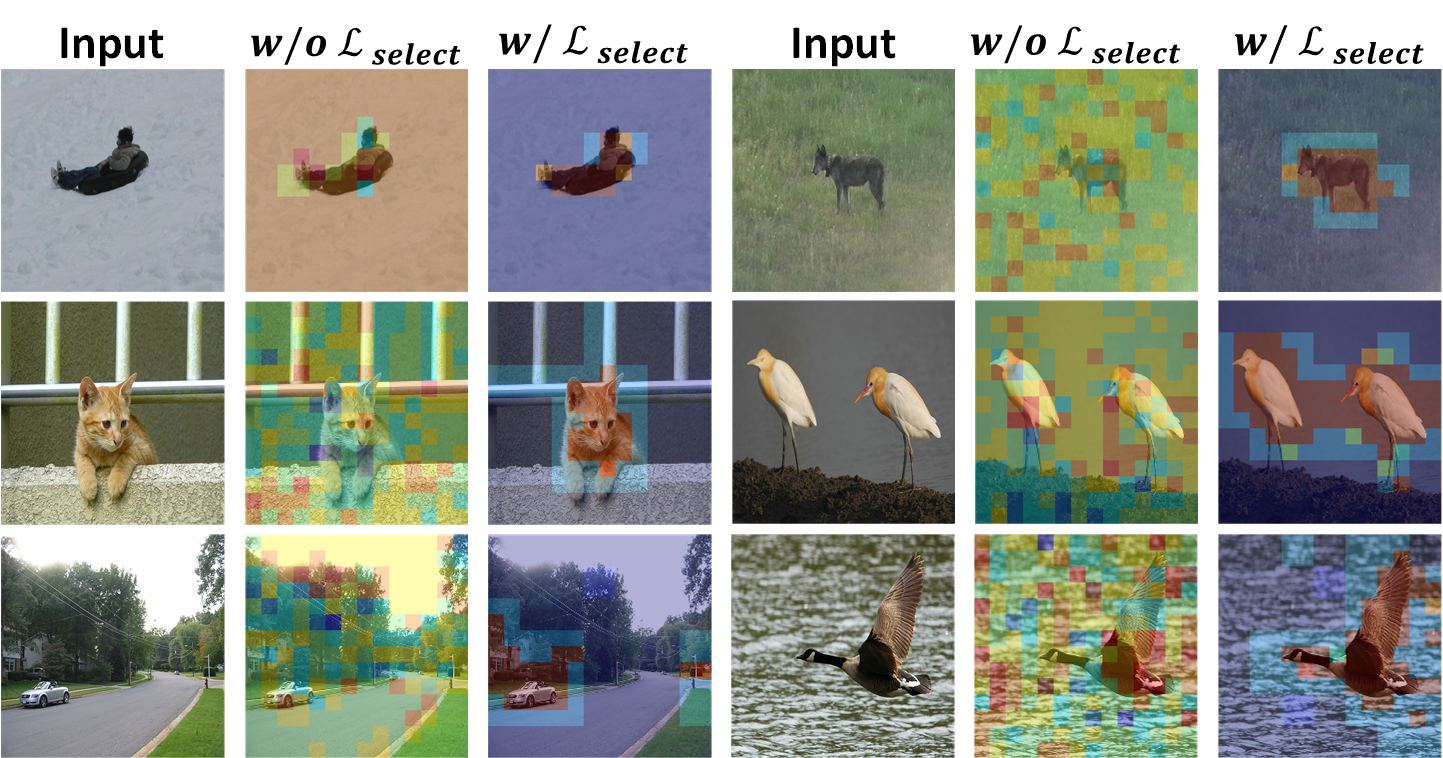}
    \caption{We overlay the summation of the selection masks generated by all SPA blocks on the original image. Warm color denotes high frequency of selection while cold color means be pruned before the attention computation. With the supervision of multi-scale select labels, the selection process becomes significantly more accurate.}
    \label{fig:ex}
\end{figure}

\subsection{Ablation Study}
\textbf{The Effect of Token Selection for Attention.}
To illustrate the effectiveness of informative token selection, we designed experiments where all informative tokens were selected based on ground truth (GT) selection. As illustrated in \cref{fig:gt}, for both plain ViT and window-based attention mechanisms, selecting tokens to disregard background information improves both accuracy and efficiency, as confirmed across two different tasks.

\textbf{Specific Design for Selection.}
Even though we know that token selection works, a critical challenge is how to correctly select these informative tokens without ground truth labels. As discussed earlier, previous methods (\eg SparseViT) commonly adopt uniform token selection, applying a fixed ratio for all images in each batch. However, the results in \cref{tab:select} demonstrate that our SPA with dynamic selection performs better. Additionally, \cref{fig:ex} provides both numeric and visual comparisons to illustrate the effectiveness of our proposed multi-scale select label.


\begin{table}[h]
\centering
\begin{tabular}{@{}c| c| c |c@{}}
\toprule
 \textbf{SPA} & $\mathcal{L}_{select}$ & \textbf{Mean Select Ratio(\%)}  &  \textbf{mAP} 
                  \\ \midrule 
                  \xmark & \xmark &  50   & 44.42    \\
     \cmark   &   \xmark &  59.77 & 44.49    \\
     \cmark   & \cmark &  \textbf{29.60 } & \textbf{44.60}    \\
   
    \bottomrule
\end{tabular}
\caption{ For uniform sparse attention, we adopt the top-50 technique as SparseViT. The results in the first row show that this method can also improve mAP (44.36 for Swin). However, our SPA block achieves better performance.}
\label{tab:select}
\end{table}

\textbf{Number of SPA blocks.}
\cref{tab:blks} explore the optimal number of SPA blocks in SPT. The results match with the findings in \cite{xia2022vision}. Starting from the third stage yields the best performance. Early-stage selection leads to information loss, resulting in worse performance.


\section{Conclusion}
In this paper, we analyze the current issues with sparse attention mechanisms and propose a novel Select and Pack (SPA) mechanism to address these challenges for both efficiency and performance. Our SPA focuses attention computations solely on informative tokens using a supervised gating block in Vision Transformers. This mechanism packs the selected tokens for parallelized GPU batch training and inference. Integrated into the Swin Transformer’s hierarchical architecture, SPA forms the efficient Select and Pack Transformer (SPT), which works as image backbone network for various computer vision tasks and generates multi-scale representations. To enhance selection accuracy and ensure effectiveness in complex computer vision tasks, we employ multi-scale selection labels for explicit supervision using object labels. Extensive experiments across four datasets and a range of vision tasks validate the effectiveness of SPT. For object detection, SPT achieves a 0.6 mAP improvement and a 16.4\% reduction in computational cost compared to state-of-the-art sparse attention mechanisms. Additionally, SPT outperforms baselines in other computer vision tasks, with a 0.24 mAP improvement in multi-label classification and a 7.05 increase in top-1 accuracy for image classification.

\bibliography{aaai25}

\begin{thebibliography}{40}
\providecommand{\natexlab}[1]{#1}

\bibitem[{Aoki, Tung, and Oliveira(2022)}]{aoki2022heterogeneous}
Aoki, R.; Tung, F.; and Oliveira, G.~L. 2022.
\newblock Heterogeneous multi-task learning with expert diversity.
\newblock \emph{IEEE/ACM Transactions on Computational Biology and Bioinformatics}, 19(6): 3093--3102.

\bibitem[{Bao et~al.(2021)Bao, Dong, Piao, and Wei}]{bao2021beit}
Bao, H.; Dong, L.; Piao, S.; and Wei, F. 2021.
\newblock Beit: Bert pre-training of image transformers.
\newblock \emph{arXiv preprint arXiv:2106.08254}.

\bibitem[{Carion et~al.(2020)Carion, Massa, Synnaeve, Usunier, Kirillov, and Zagoruyko}]{carion2020end}
Carion, N.; Massa, F.; Synnaeve, G.; Usunier, N.; Kirillov, A.; and Zagoruyko, S. 2020.
\newblock End-to-end object detection with transformers.
\newblock In \emph{European conference on computer vision}, 213--229. Springer.

\bibitem[{Caron et~al.(2021)Caron, Touvron, Misra, J{\'e}gou, Mairal, Bojanowski, and Joulin}]{caron2021emerging}
Caron, M.; Touvron, H.; Misra, I.; J{\'e}gou, H.; Mairal, J.; Bojanowski, P.; and Joulin, A. 2021.
\newblock Emerging properties in self-supervised vision transformers.
\newblock In \emph{Proceedings of the IEEE/CVF international conference on computer vision}, 9650--9660.

\bibitem[{Chen et~al.(2023)Chen, Liu, Tang, Yi, Zhao, and Han}]{chen2023sparsevit}
Chen, X.; Liu, Z.; Tang, H.; Yi, L.; Zhao, H.; and Han, S. 2023.
\newblock SparseViT: Revisiting Activation Sparsity for Efficient High-Resolution Vision Transformer.
\newblock In \emph{Proceedings of the IEEE/CVF Conference on Computer Vision and Pattern Recognition}, 2061--2070.

\bibitem[{Chen et~al.(2022)Chen, Deng, Wu, Gu, and Li}]{chen2022towards}
Chen, Z.; Deng, Y.; Wu, Y.; Gu, Q.; and Li, Y. 2022.
\newblock Towards understanding mixture of experts in deep learning.
\newblock \emph{arXiv preprint arXiv:2208.02813}.

\bibitem[{Child et~al.(2019)Child, Gray, Radford, and Sutskever}]{child2019generating}
Child, R.; Gray, S.; Radford, A.; and Sutskever, I. 2019.
\newblock Generating long sequences with sparse transformers.
\newblock \emph{arXiv preprint arXiv:1904.10509}.

\bibitem[{Dai et~al.(2017)Dai, Qi, Xiong, Li, Zhang, Hu, and Wei}]{dai2017deformable}
Dai, J.; Qi, H.; Xiong, Y.; Li, Y.; Zhang, G.; Hu, H.; and Wei, Y. 2017.
\newblock Deformable convolutional networks.
\newblock In \emph{Proceedings of the IEEE international conference on computer vision}, 764--773.

\bibitem[{Dehghani et~al.(2024)Dehghani, Mustafa, Djolonga, Heek, Minderer, Caron, Steiner, Puigcerver, Geirhos, Alabdulmohsin et~al.}]{dehghani2024patch}
Dehghani, M.; Mustafa, B.; Djolonga, J.; Heek, J.; Minderer, M.; Caron, M.; Steiner, A.; Puigcerver, J.; Geirhos, R.; Alabdulmohsin, I.~M.; et~al. 2024.
\newblock Patch n’pack: Navit, a vision transformer for any aspect ratio and resolution.
\newblock \emph{Advances in Neural Information Processing Systems}, 36.

\bibitem[{Dosovitskiy et~al.(2020)Dosovitskiy, Beyer, Kolesnikov, Weissenborn, Zhai, Unterthiner, Dehghani, Minderer, Heigold, Gelly et~al.}]{dosovitskiy2020image}
Dosovitskiy, A.; Beyer, L.; Kolesnikov, A.; Weissenborn, D.; Zhai, X.; Unterthiner, T.; Dehghani, M.; Minderer, M.; Heigold, G.; Gelly, S.; et~al. 2020.
\newblock An image is worth 16x16 words: Transformers for image recognition at scale.
\newblock \emph{arXiv preprint arXiv:2010.11929}.

\bibitem[{Fan and Tao(2024)}]{fan2024towards}
Fan, X.; and Tao, C. 2024.
\newblock Towards Resilient and Efficient LLMs: A Comparative Study of Efficiency, Performance, and Adversarial Robustness.
\newblock \emph{arXiv preprint arXiv:2408.04585}.

\bibitem[{Fan, Tao, and Zhao(2024)}]{fan2024advanced}
Fan, X.; Tao, C.; and Zhao, J. 2024.
\newblock Advanced Stock Price Prediction with xLSTM-Based Models: Improving Long-Term Forecasting.
\newblock \emph{Preprints}, (2024082109).

\bibitem[{He et~al.(2022)He, Chen, Xie, Li, Doll{\'a}r, and Girshick}]{he2022masked}
He, K.; Chen, X.; Xie, S.; Li, Y.; Doll{\'a}r, P.; and Girshick, R. 2022.
\newblock Masked autoencoders are scalable vision learners.
\newblock In \emph{Proceedings of the IEEE/CVF conference on computer vision and pattern recognition}, 16000--16009.

\bibitem[{Hua et~al.(2022)Hua, Dai, Liu, and Le}]{hua2022transformer}
Hua, W.; Dai, Z.; Liu, H.; and Le, Q. 2022.
\newblock Transformer quality in linear time.
\newblock In \emph{International conference on machine learning}, 9099--9117. PMLR.

\bibitem[{Huang et~al.(2020)Huang, She, Wang, and Zhang}]{huang2020gatenet}
Huang, T.; She, Q.; Wang, Z.; and Zhang, J. 2020.
\newblock GateNet: gating-enhanced deep network for click-through rate prediction.
\newblock \emph{arXiv preprint arXiv:2007.03519}.

\bibitem[{Jang, Gu, and Poole(2016)}]{jang2016categorical}
Jang, E.; Gu, S.; and Poole, B. 2016.
\newblock Categorical reparameterization with gumbel-softmax.
\newblock \emph{arXiv preprint arXiv:1611.01144}.

\bibitem[{Kang et~al.(2022)Kang, Zhang, Zhao, Yang, and Yang}]{kang2022tie}
Kang, Y.; Zhang, Z.; Zhao, M.; Yang, X.; and Yang, X. 2022.
\newblock Tie Memories to E-souvenirs: Hybrid Tangible AR Souvenirs in the Museum.
\newblock In \emph{Adjunct Proceedings of the 35th Annual ACM Symposium on User Interface Software and Technology}, 1--3.

\bibitem[{Kirillov et~al.(2023)Kirillov, Mintun, Ravi, Mao, Rolland, Gustafson, Xiao, Whitehead, Berg, Lo et~al.}]{kirillov2023segment}
Kirillov, A.; Mintun, E.; Ravi, N.; Mao, H.; Rolland, C.; Gustafson, L.; Xiao, T.; Whitehead, S.; Berg, A.~C.; Lo, W.-Y.; et~al. 2023.
\newblock Segment anything.
\newblock \emph{arXiv preprint arXiv:2304.02643}.

\bibitem[{Konstantinidis et~al.(2023)Konstantinidis, Papastratis, Dimitropoulos, and Daras}]{konstantinidis2023multi}
Konstantinidis, D.; Papastratis, I.; Dimitropoulos, K.; and Daras, P. 2023.
\newblock Multi-manifold attention for vision transformers.
\newblock \emph{IEEE Access}.

\bibitem[{Liang et~al.(2021)Liang, Cao, Sun, Zhang, Van~Gool, and Timofte}]{liang2021swinir}
Liang, J.; Cao, J.; Sun, G.; Zhang, K.; Van~Gool, L.; and Timofte, R. 2021.
\newblock Swinir: Image restoration using swin transformer.
\newblock In \emph{Proceedings of the IEEE/CVF international conference on computer vision}, 1833--1844.

\bibitem[{Liang et~al.(2022)Liang, Ge, Tong, Song, Wang, and Xie}]{liang2022not}
Liang, Y.; Ge, C.; Tong, Z.; Song, Y.; Wang, J.; and Xie, P. 2022.
\newblock Not all patches are what you need: Expediting vision transformers via token reorganizations.
\newblock \emph{arXiv preprint arXiv:2202.07800}.

\bibitem[{Lin et~al.(2021)Lin, Yang, He, Fan, and Zhang}]{lin2021metagater}
Lin, S.; Yang, L.; He, Z.; Fan, D.; and Zhang, J. 2021.
\newblock MetaGater: Fast learning of conditional channel gated networks via federated meta-learning.
\newblock In \emph{2021 IEEE 18th International Conference on Mobile Ad Hoc and Smart Systems (MASS)}, 164--172. IEEE.

\bibitem[{Lingle(2023)}]{lingle2023transformer}
Lingle, L.~D. 2023.
\newblock Transformer-vq: Linear-time transformers via vector quantization.
\newblock \emph{arXiv preprint arXiv:2309.16354}.

\bibitem[{Liu et~al.(2022)Liu, Hu, Lin, Yao, Xie, Wei, Ning, Cao, Zhang, Dong et~al.}]{liu2022swin}
Liu, Z.; Hu, H.; Lin, Y.; Yao, Z.; Xie, Z.; Wei, Y.; Ning, J.; Cao, Y.; Zhang, Z.; Dong, L.; et~al. 2022.
\newblock Swin transformer v2: Scaling up capacity and resolution.
\newblock In \emph{Proceedings of the IEEE/CVF conference on computer vision and pattern recognition}, 12009--12019.

\bibitem[{Liu et~al.(2021)Liu, Lin, Cao, Hu, Wei, Zhang, Lin, and Guo}]{liu2021swin}
Liu, Z.; Lin, Y.; Cao, Y.; Hu, H.; Wei, Y.; Zhang, Z.; Lin, S.; and Guo, B. 2021.
\newblock Swin transformer: Hierarchical vision transformer using shifted windows.
\newblock In \emph{Proceedings of the IEEE/CVF international conference on computer vision}, 10012--10022.

\bibitem[{Ni et~al.(2024{\natexlab{a}})Ni, Meng, Chen, Zhao, Chen, Li, Zhang, Yin, Wang, and Chan}]{ni2024earnings}
Ni, H.; Meng, S.; Chen, X.; Zhao, Z.; Chen, A.; Li, P.; Zhang, S.; Yin, Q.; Wang, Y.; and Chan, Y. 2024{\natexlab{a}}.
\newblock Harnessing Earnings Reports for Stock Predictions: A QLoRA-Enhanced LLM Approach.
\newblock \emph{arXiv preprint arXiv:2408.06634}.

\bibitem[{Ni et~al.(2024{\natexlab{b}})Ni, Meng, Geng, Li, Li, Chen, Wang, and Zhang}]{ni2024timeseries}
Ni, H.; Meng, S.; Geng, X.; Li, P.; Li, Z.; Chen, X.; Wang, X.; and Zhang, S. 2024{\natexlab{b}}.
\newblock Time Series Modeling for Heart Rate Prediction: From ARIMA to Transformers.
\newblock \emph{arXiv preprint arXiv:2406.12199}.

\bibitem[{Petersen et~al.(2022)Petersen, Borgelt, Kuehne, and Deussen}]{petersen2022deep}
Petersen, F.; Borgelt, C.; Kuehne, H.; and Deussen, O. 2022.
\newblock Deep differentiable logic gate networks.
\newblock \emph{Advances in Neural Information Processing Systems}, 35: 2006--2018.

\bibitem[{Rao et~al.(2023)Rao, Liu, Zhao, Zhou, and Lu}]{rao2023dynamic}
Rao, Y.; Liu, Z.; Zhao, W.; Zhou, J.; and Lu, J. 2023.
\newblock Dynamic spatial sparsification for efficient vision transformers and convolutional neural networks.
\newblock \emph{IEEE Transactions on Pattern Analysis and Machine Intelligence}, 45(9): 10883--10897.

\bibitem[{Rao et~al.(2021)Rao, Zhao, Liu, Lu, Zhou, and Hsieh}]{rao2021dynamicvit}
Rao, Y.; Zhao, W.; Liu, B.; Lu, J.; Zhou, J.; and Hsieh, C.-J. 2021.
\newblock Dynamicvit: Efficient vision transformers with dynamic token sparsification.
\newblock \emph{Advances in neural information processing systems}, 34: 13937--13949.

\bibitem[{Shazeer(2020)}]{shazeer2020glu}
Shazeer, N. 2020.
\newblock Glu variants improve transformer.
\newblock \emph{arXiv preprint arXiv:2002.05202}.

\bibitem[{Shazeer et~al.(2017)Shazeer, Mirhoseini, Maziarz, Davis, Le, Hinton, and Dean}]{shazeer2017outrageously}
Shazeer, N.; Mirhoseini, A.; Maziarz, K.; Davis, A.; Le, Q.; Hinton, G.; and Dean, J. 2017.
\newblock Outrageously large neural networks: The sparsely-gated mixture-of-experts layer.
\newblock \emph{arXiv preprint arXiv:1701.06538}.

\bibitem[{Touvron et~al.(2021)Touvron, Cord, Douze, Massa, Sablayrolles, and J{\'e}gou}]{touvron2021training}
Touvron, H.; Cord, M.; Douze, M.; Massa, F.; Sablayrolles, A.; and J{\'e}gou, H. 2021.
\newblock Training data-efficient image transformers \& distillation through attention.
\newblock In \emph{International conference on machine learning}, 10347--10357. PMLR.

\bibitem[{Vaswani et~al.(2017)Vaswani, Shazeer, Parmar, Uszkoreit, Jones, Gomez, Kaiser, and Polosukhin}]{vaswani2017attention}
Vaswani, A.; Shazeer, N.; Parmar, N.; Uszkoreit, J.; Jones, L.; Gomez, A.~N.; Kaiser, {\L}.; and Polosukhin, I. 2017.
\newblock Attention is all you need.
\newblock \emph{Advances in neural information processing systems}, 30.

\bibitem[{Xia et~al.(2022)Xia, Pan, Song, Li, and Huang}]{xia2022vision}
Xia, Z.; Pan, X.; Song, S.; Li, L.~E.; and Huang, G. 2022.
\newblock Vision transformer with deformable attention.
\newblock In \emph{Proceedings of the IEEE/CVF conference on computer vision and pattern recognition}, 4794--4803.

\bibitem[{Yang, Kang, and Yang(2022)}]{yang2022retargeting}
Yang, X.; Kang, Y.; and Yang, X. 2022.
\newblock Retargeting destinations of passive props for enhancing haptic feedback in virtual reality.
\newblock In \emph{2022 IEEE Conference on Virtual Reality and 3D User Interfaces Abstracts and Workshops (VRW)}, 618--619. IEEE.

\bibitem[{Ye et~al.(2023)Ye, Fang, Du, Yuen, and Tao}]{ye2023heterogeneous}
Ye, M.; Fang, X.; Du, B.; Yuen, P.~C.; and Tao, D. 2023.
\newblock Heterogeneous federated learning: State-of-the-art and research challenges.
\newblock \emph{ACM Computing Surveys}, 56(3): 1--44.

\bibitem[{Zhang et~al.(2023)Zhang, Kasichainula, Zhuo, Li, Seo, and Cao}]{zhang2023transformer}
Zhang, T.; Kasichainula, K.; Zhuo, Y.; Li, B.; Seo, J.-s.; and Cao, Y. 2023.
\newblock Transformer-based Selective Super-Resolution for Efficient Image Refinement.
\newblock \emph{arXiv preprint arXiv:2312.05803}.

\bibitem[{Zhang et~al.(2024)Zhang, Kasichainula, Zhuo, Li, Seo, and Cao}]{zhang2024patch}
Zhang, T.; Kasichainula, K.; Zhuo, Y.; Li, B.; Seo, J.-S.; and Cao, Y. 2024.
\newblock Patch-based Selection and Refinement for Early Object Detection.
\newblock In \emph{Proceedings of the IEEE/CVF Winter Conference on Applications of Computer Vision}, 729--738.

\bibitem[{Zhou et~al.(2024)Zhou, Zeng, Chen, Zhou, Ni, Zhang, Li, Liu, Zheng, and Chen}]{zhou2024reconstruction}
Zhou, Y.; Zeng, Z.; Chen, A.; Zhou, X.; Ni, H.; Zhang, S.; Li, P.; Liu, L.; Zheng, M.; and Chen, X. 2024.
\newblock Evaluating Modern Approaches in 3D Scene Reconstruction: NeRF vs Gaussian-Based Methods.
\newblock \emph{arXiv preprint arXiv:2408.04268}.

\end{thebibliography}

\end{document}